\documentclass{article}
\usepackage{ijcai25}

\usepackage{times}
\usepackage{soul}
\usepackage{url}
\usepackage[hidelinks]{hyperref}
\usepackage[utf8]{inputenc}
\usepackage[small]{caption}
\usepackage{graphicx}
\usepackage{amsmath,amssymb,amsfonts}
\usepackage{amsthm}
\usepackage{booktabs}
\usepackage{algorithm}
\usepackage{algorithmic}
\usepackage[switch]{lineno}
\usepackage{enumitem}
\usepackage{placeins}

\title{Reinforcement Learning for Hybrid Charging Stations Planning and Operation Considering Fixed and Mobile Chargers}

\author{
Yanchen Zhu$^1$
\and
Honghui Zou$^2$\and
Chufan Liu$^{1}$\and
Yuyu Luo$^1$\and
Yuankai Wu$^3$\And
Yuxuan Liang$^{1,}$\thanks{Corresponding author. Email: yuxliang@outlook.com}
\\
\affiliations
$^1$Hong Kong University of Science and Technology (Guangzhou)\\
$^2$Beihang University\\
$^3$Sichuan University\\
\emails
yzhu156@connect.hkust-gz.edu.cn,
hhzou@buaa.edu.cn,
cliu779@connect.hkust-gz.edu.cn,
yuyuluo@hkust-gz.edu.cn,
Kaimaogege@gmail.com,
yuxliang@outlook.com
}
\begin{document}

\maketitle

\begin{abstract}

The success of vehicle electrification relies on efficient and adaptable charging infrastructure. Fixed-location charging stations often suffer from underutilization or congestion due to fluctuating demand, while mobile chargers offer flexibility by relocating as needed. This paper studies the optimal planning and operation of hybrid charging infrastructures that combine both fixed and mobile chargers within urban road networks. We formulate the Hybrid Charging Station Planning and Operation (HCSPO) problem, jointly optimizing the placement of fixed stations and the scheduling of mobile chargers. A charging demand prediction model based on Model Predictive Control (MPC) supports dynamic decision-making. To solve the HCSPO problem, we propose a deep reinforcement learning approach enhanced with heuristic scheduling. Experiments on real-world urban scenarios show that our method improves infrastructure availability—achieving up to 244.4\% increase in coverage—and reduces user inconvenience with up to 79.8\% shorter waiting times, compared to existing solutions.
\end{abstract}

\section{Introduction}

The transition towards vehicle electrification is rapidly advancing, driven by its substantial societal and environmental benefits. However, the widespread adoption of electric vehicles (EVs) hinges on the availability and efficiency of charging infrastructure \cite{RN17}, which requires a comprehensive and adaptable planning approach to accommodate the increasing demand. A critical challenge in this process is accounting for the dynamic nature of charging demand, which is often overlooked in traditional station planning methods.

\begin{figure}
    \centering
    \includegraphics[width=\linewidth]{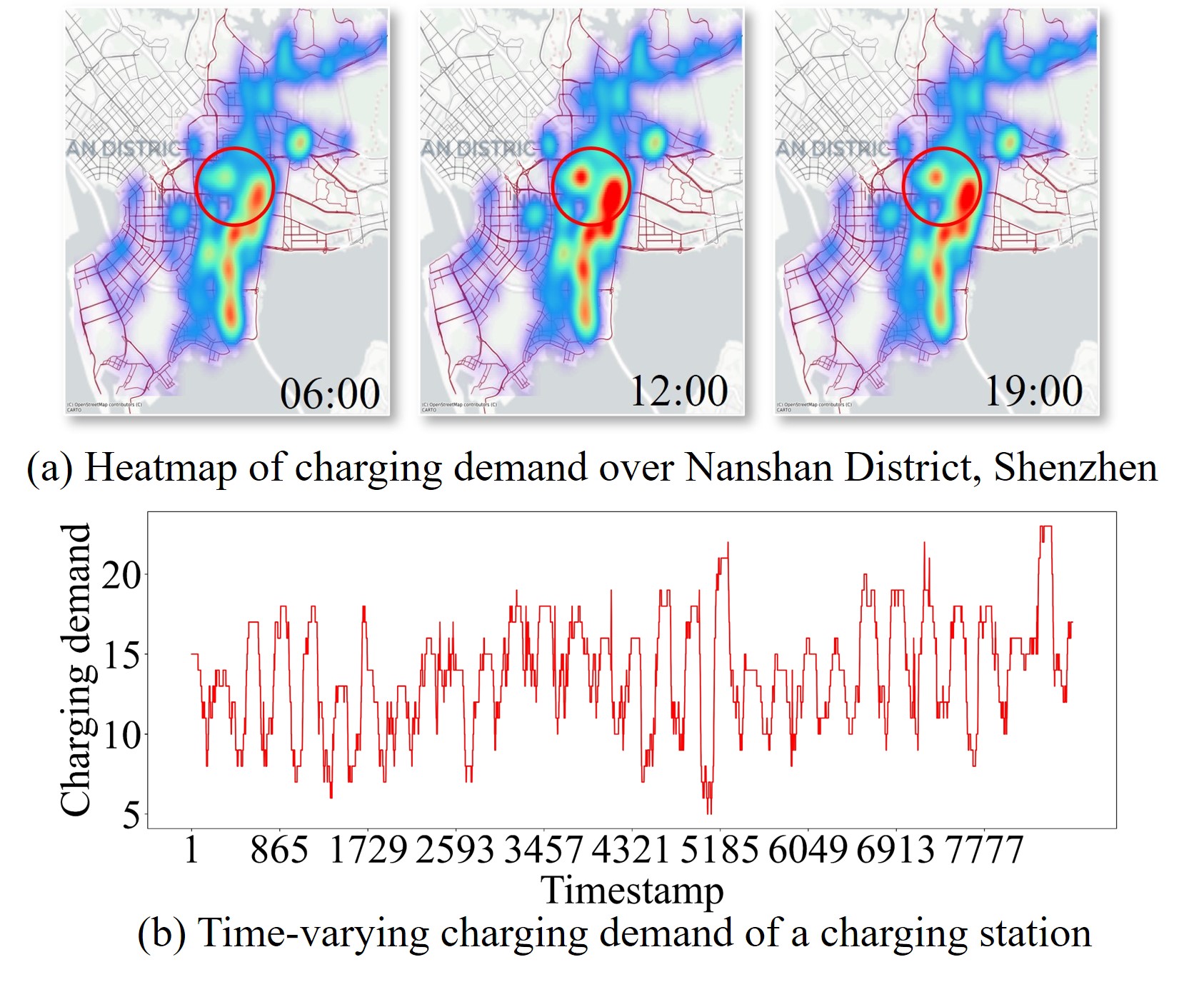}
    % \vspace{-1em}
    % \vspace{-1em}
    \caption{Dynamic attribute of charging demand}
    \label{fig:dynamic demand}
    % \vspace{-1em}

\end{figure}

As shown in Figure \ref{fig:dynamic demand}, panel (a) presents a heat map of charging demand across Nanshan District in Shenzhen, China, highlighting how demand distribution varies throughout the day, from 6:00 AM to 7:00 PM. Meanwhile, panel (b) depicts the time series of charging demand for a specific station in Shenzhen, further illustrating the fluctuations in demand at a granular level. These spatio-temporal patterns collectively emphasize the dynamic nature of charging demand in urban areas.

% While traditional charging stations, primarily consisting of fixed-location chargers, offer cost advantages and power grid stability through permanent installations, their spatial rigidity leads to systemic inefficiencies, particularly in addressing the demand-supply mismatch caused by fluctuating demand patterns. For instance, if the number of charging stations is based on peak demand periods, a significant portion of the infrastructure may remain underutilized during off-peak hours. Conversely, sizing infrastructure for off-peak demand may result in congestion and insufficient capacity during peak hours. These limitations emphasize the critical need for adaptive infrastructure solutions capable of reconfiguring in response to changing demand.

While traditional charging stations with fixed-location chargers offer cost and grid stability benefits, their spatial rigidity leads to inefficiencies, especially in addressing demand-supply mismatches from fluctuating demand. For example, sizing infrastructure for peak demand causes underutilization during off-peak hours, while sizing for off-peak leads to congestion during peaks. These limitations highlight the need for adaptive infrastructure that can reconfigure in response to changing demand.

Emerging mobile chargers (MCs) have recently gained attention as a flexible supplement to urban charging networks \cite{MCS7}. Unlike fixed chargers, mobile chargers can be dynamically scheduled to align with changing demand patterns, offering a promising solution to the limitations of traditional charging infrastructure. In addition to addressing the limitations of fixed charging stations, the integration of MCs can also benefit the configuration of fixed charging stations. For example, in areas with demand surges, MCs can be scheduled for temporary service, eliminating the need for installing fixed stations in these locations. This approach ensures that fixed stations are deployed where demand is more consistent, improving overall system efficiency.

Considering optimization for both fixed and mobile charging infrastructures, this paper introduces a novel hybrid charging station planning and operation (HCSPO) problem, which aims to integrate the optimal planning of fixed charging stations with dynamic operation of mobile chargers. Specifically, we formulate the HCSPO problem within urban road networks to optimize the locations and configurations of fixed charging stations and to dynamically schedule mobile chargers to support real-time charging needs.

\begin{figure*}[t]
    \centering
    \includegraphics[width=0.92\textwidth]{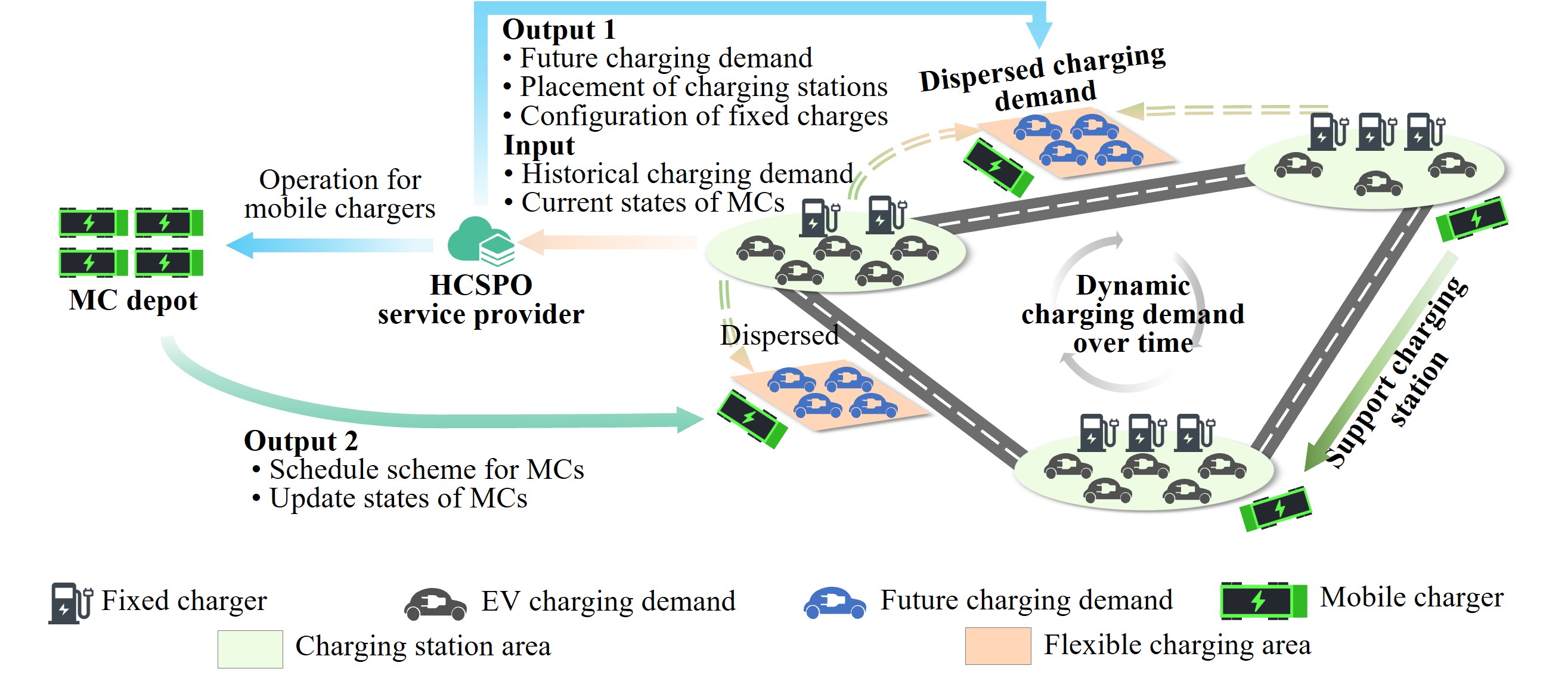}
    % \vspace{-1em}    
    % \vspace{-1em}
    \caption{Overview of HCSPO}
    \label{fig:Overview of HCSPO}
    % \vspace{-1em}
\end{figure*}

To enhance decision-making, we incorporate a charging demand prediction model based on Model Predictive Control (MPC) to provide foresight into demand fluctuations. Our solution leverages a reinforcement learning (RL) method, augmented with heuristic techniques, to adaptively optimize the planning and operation of the HCSPO problem. This integrated approach effectively bridges the strategic planning of fixed charging stations with the operational dynamics of mobile chargers, contributing to the sustainability and accessibility of urban transportation networks.

In summary, the contributions of this paper are as follows: 
\begin{itemize}[leftmargin=*]
    \item We formulate the HCSPO problem into a rolling horizon framework within road networks, integrating the planning of fixed stations with mobile charging operations. Our approach incorporates a demand prediction model based on an MPC policy to enhance decision-making. 
    \item We propose an adaptive RL algorithm, enhanced with heuristic scheduling techniques, to efficiently tackle the HCSPO problem, ensuring optimal planning and operation of both fixed and mobile infrastructures. 
    \item We conduct extensive experiments on real-world dataset, which demonstrates our approach outperforms other baselines, offering superior societal benefits, including improved sustainability and user satisfaction. 
\end{itemize}

\section{Related Work}

This section reviews related work on charging station planning and mobile charging infrastructure operations.

EV charging station planning is often modeled as a variant of the Facility Location Problem. However, many approaches rely on simplified assumptions about charging demand \cite{RN1}, which may not reflect actual spatio-temporal distributions. Some studies simulate EV recharging behavior using real-world datasets like GPS records \cite{RN2,RN3,RN4,RN5,RN6}, while others, such as \cite{RN3,RN7}, use taxi trajectories. Recent research also explores dynamic and stochastic environments \cite{RN8,RN9,RN10,RN11,RN12,RN13,RN14}. 

Mobile charging station placement is gaining traction. For example, \cite{MCS3} proposes a two-phase framework to determine MCS placement, scheduling, and relocation, which is widely adopted in MCS placement and scheduling studies, including depot and fleet locations \cite{MCS1,MCS2,MCS4,MCS5,MCS6}. Some studies validate approaches with real-world data, including GPS trajectories \cite{MCS1,MCS2}. \cite{MCS2} uses a Markov Decision Process (MDP) to reduce delays and increase the proportion of charged EVs, while \cite{MCS1} extends the two-phase method with a multi-agent RL algorithm for dynamic MCS operation.

While substantial research exists on both fixed-location and mobile charging infrastructures, to our best knowledge, no study has integrated these within a single framework. This paper proposes an adaptive RL approach that simultaneously determines fixed station locations and configurations while dynamically scheduling mobile chargers.

\section{Problem Statement}

In this section, we formally define the HCSPO problem and provide the following key definitions:

\newtheorem{definition}{Definition}
\begin{definition}[Road Network]
Let $G = (V, E)$ be a directed weighted graph with $V$ the set of vertices and $E$ the set of edges. The vertices are the road network junctions, while the edges represent the roads direction-wise.
\end{definition}

% Based on road network, we define charging demand of each vertices and how it varies over time.

\begin{definition}[Dynamic Charging Demand]
Considering the time-varied nature of charging demand, Let the recording $dem^{1,\dots,|T|}(v)=\left[dem^1(v),dem^2(v),\cdots, dem^{|T|}(v)\right]$ denote the charging demand of vertex $v$ over the operation horizon divided into $|T|$ time slots.
\end{definition}

% Next we introduce charging stations as well as mobile chargers located within the road network $G$.

\begin{definition}[Charging Station]
A charging station (CS) $s$ within the road network $G$ is defined as a tuple $s = (v, x)$, where $v \in V$ is the location node, $x = (x_1, \ldots, x_n)$ is a vector of length $n \in \mathbb{N}$ with $x_i \in \mathbb{N}$ being the number of fixed chargers of type $i$ at $s$. We denote the set of all possible CSs as $S$. We set a limit on the number of fixed chargers of each $s$ by $\sum_ix_i \leq K$. 
\end{definition}

\begin{definition}[Mobile Charger and Depot]
A mobile charger (MC) $m$ can be defined as $m=(l^{1,\cdots,|T|}_m,\tau^{1,\cdots,|T|}_m,q^{1,\cdots,|T|}_m)$, which denotes the current location, accessible time (arrival time after scheduling decision) and remaining electricity of $m$ at different time slots $t\in T$. Each schedule for MCs is batched with size of $k_{MC}$ as a fleet. In addition, we define mobile depots $j \in J$ as the initial locations of MCs before scheduling and where they return to recharge when their batteries deplete, i.e., $l^1_m = j \in J$.
\end{definition}

\begin{definition}[Charging Plan]
A charging plan $P=(S,M)$ on $G$ includes an assignment of vertices to stations $s \in S$ including the configuration of fixed chargers and the scheduling of each mobile charger among all time slots. 
\end{definition}

We use a utility function (see Eq. (\ref{eq: utility})) to evaluate the effectiveness of a charging plan. We set a limit on the budget to install fixed stations via Eq. \ref{eq: budget limitation}.
\begin{equation}
    \label{eq: budget limitation}
    \sum_{s\in S}{x_i\cdot fee_i}+\sum_{m\in M}{i_{MC}(m)\cdot fee_{MC}}\leq Budget.
\end{equation}

Here, $fee_i$ represents the installation cost of the $i$-th type of fixed charger, $i_{MC}(m)$ is a binary variable where $i_{MC}(m) = 1$ indicates that mobile charger $m$ is employed; otherwise, it remains idle throughout the entire horizon. $fee_{MC}$ denotes the operational cost of each mobile charger, and $Budget$ refers to the total budget allocated for the system.

We formalize HCSPO as follows. Given a road network\ $G=(V,E)$, time slots\ $T={1,\dots,|T|}$, historical node‑level demand, and perfect foresight of future demand, we seek a charging plan that satisfies all demand and maximizes a utility that balances social benefit against installation cost and queuing loss.

Figure~\ref{fig:Overview of HCSPO} summarizes the workflow: before operation, the provider forecasts demand from historical data, selects station locations/configurations, and schedules mobile chargers to flexibly serve areas beyond fixed stations and ease peak‑hour congestion.

\section{System Framework}

\begin{figure}[t]
    \centering
    \includegraphics[width=\linewidth]{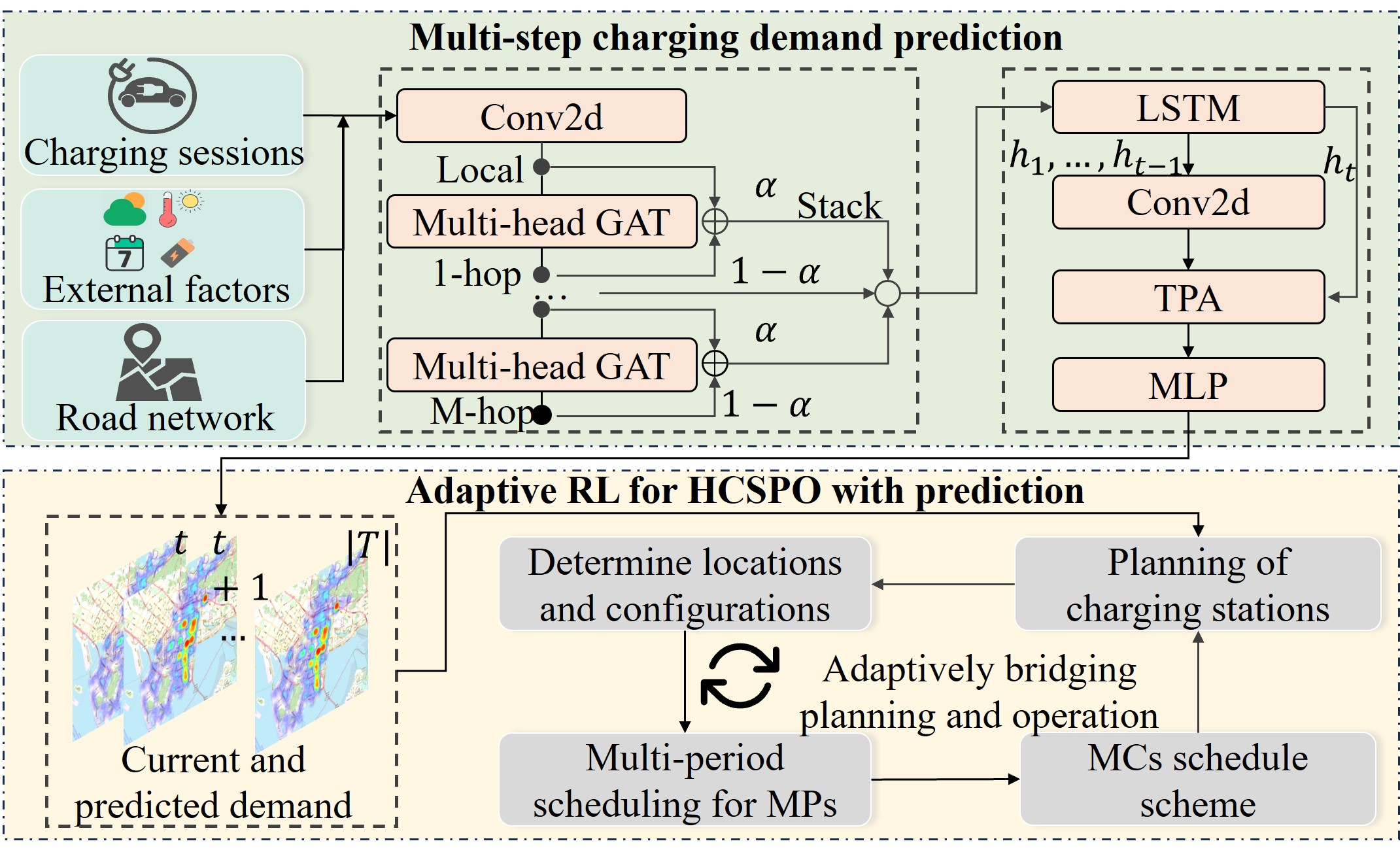}
    % \vspace{-1em}    
    % \vspace{-1em}
    \caption{System framework}
    \label{fig:System framework}
    % \vspace{-1em}
\end{figure}

This section outlines the HCSPO system framework (Figure~\ref{fig:System framework}). We begin with a multi-step charging demand prediction model based on historical data. Next, we adapt the utility model from \cite{RN17} into a rolling horizon framework that incorporates current and future demand for MPC-based decision-making. Finally, we propose two mobile charger scheduling strategies integrated with fixed station planning.

% \subsection{Hybrid Charging Station Planning and Operation}
% sadsadsad

\subsection{Multi-step Charging Demand Prediction}
\label{sec: demand prediction}
To anticipate future charging needs, we formulate a multi-step demand prediction problem as:
\begin{equation}
\hat{dem}^{t+1:t+\omega_{pred}} = f(dem^{t-\omega_{hist}+1:t}, F^{t-\omega_{hist}+1:t}),
\end{equation}
where $dem$ denotes historical node-level demand and $F$ includes exogenous features (e.g., weather, day type). We adopt a spatiotemporal forecasting model inspired by \cite{RN16}, which integrates a \textbf{Graph Embedding Module} and a \textbf{Multivariate Decoder}. The embedding module captures temporal trends via CNN and spatial correlations via GAT~\cite{GAT}, enhanced by residual connections. The decoder uses TPA-LSTM~\cite{TPALSTM} to extract long-term temporal dependencies with multi-hop attention, enabling accurate multi-step forecasting over the graph. This predicted demand guides downstream optimization of CS and MC operations.

\subsection{Utility Function}
\label{sec: utility model}

This section introduces a utility model to evaluate the effectiveness of a charging plan. Building on the utility model designed by \cite{RN17}, we extend this model into a rolling horizon framework to assess the performance of a charging plan under time-varying demand. We also introduce a queuing loss metric that measures the number of EVs leaving the queue due to excessive wait times for recharging.

\subsubsection{Benefit}

The benefit function measures the coverage of the road network by charging infrastructures. Intuitively, nodes supported by a greater number of charging infrastructures suggest higher benefits. To quantify this, we define the influential radius of a charging station (CS) $s$ as the distance within which it attracts EVs from nearby nodes, influenced by its charging supply capacity.\par

We first define the charging supply capacity as the total power provided by both fixed chargers and mobile chargers (if any) at time slot $t$, which can be calculated as follows:
\begin{equation}
C^t(s) = \sum_{i=1}^n x_i c_i + \sum_{m \in M} i_{MC}(m,s,t)c_{MC}k_{MC}\delta,
\end{equation}
where $c_i$ denotes the charging power provided by each fixed charger of type $i$, $c_{MC}$ is the charging power provided by each mobile charger (MC), and $i_{MC}(m,s,t): M \times V \times T \mapsto \{0,1\}$ is an indicator function that equals 1 if $m$ is scheduled to $s$ at time slot $t$, and 0 otherwise. $\delta \in [0,1]$ is the invalid time discount factor since MCs could not offer charging capacity while scheduling, which will be described in Section \ref{sec: energy adjustment}. Once the capacity of a CS $s$ is determined, the influential radius of $s$ at time slot $t$ is: $R^t(s) = R_{\text{max}} \frac{1}{1 + \exp\left(-\tilde{C}^t(s)\right)}$, 
% \begin{equation}
% R^t(s) = R_{\text{max}} \frac{1}{1 + \exp\left(-\tilde{C}^t(s)\right)},
% \end{equation}
where $R_{\text{max}}$ denotes the maximum influential radius (a model parameter), and $-\tilde{C}^t(s)$ represents the scaled capacity of $s$ at $t$.\par

Next, we define the coverage of node $v$ at $t$, $cov^t(v)$, as the set of stations whose influential radius can cover $v$: $cov^t(v) = \{s \in P \mid dist(v,s) < R^t(s)\}$, where $dist(v,s)$ is the distance between $v$ and $s$. The benefit of the charging plan $P$ is then formulated as follows:
\begin{equation}
benefit(P) = \frac{1}{|V||T|} \sum_{t \in T} \sum_{v \in V} \left( \sum_{i=1}^{|cov^t(v)|} \frac{1}{i} \right).
\end{equation}

\subsubsection{Cost}
\label{sec: cost funtion}
We evaluate the cost of a charging plan $P$ in terms of $travel\ time$, $charging\ time$ and $waiting\ time$. However, since the presence of dynamic charging demand and MCs scheduling, EVs from the same node may visit different CSs at different time slots. Thus we also need to extend cost function into a rolling horizon context.
In order to estimate $travel\ time$, we first define an indicator function $i_{CS}(v,s): V \times S \mapsto \{0,1\}$ to be 1 if $s$ is the CS assigned to $v$ according to Station Seeking Algorithm in \cite{RN18}, and 0 otherwise. Then the $travel\ time$ of $P$ is:
\begin{equation}
travel(P)=\frac{1}{|T|}\sum_{t\in T}\sum_{v\in V}\sum_{s\in S}i_{CS}(v,s)\frac{dist(v,s)}{vel}dem^t(v),
\end{equation}
where $vel=const$ is the average vehicle speed. \par

Next, we estimate the $charging\ time$ and $waiting\ time$ using the Pollaczek-Khintchine formula, as in \cite{RN17}. However, this approach assumes no limits on queue length, considering only the stability of the queuing system. In real-world applications, charging demand can vary over time, and demand may occasionally exceed station capacity. To address this, instead of enforcing stability constraints, we set a maximum average waiting time, $W_{max}$. Specifically, if the waiting time for a queue with length $s$ exceeds $W_{\text{max}}$, no new EV will join the queue. Let

% \begin{equation}
%     \mu^t(s)=\frac{C^t(s)}{B}
%     \label{mu}
% \end{equation}

\begin{equation}
    \mu^t(s)={C^t(s)}/{B}
    \label{mu}
\end{equation}

be the service rate of $s$ at $t$ (where $B=const$ is the electricity required for recharging each EV). According to Pollaczek-Khintchine formula, estimated number of approaching EVs at $t$ $D^t(s)$ is:
\begin{equation}
D^t(s) = \sum_{v \in V} \frac{i_{CS}(v,s)}{dist(v,s)}dem^t(v). 
\end{equation}
Then the average waiting time of station $s$ with no stability constraints is:
\begin{equation}
W^t(s)=\frac{ \rho ^t(s)}{2 \mu ^t(s)\left(1- \rho ^t(s)\right)},
\label{W(s)}
\end{equation}
where 
\begin{equation}
\rho ^t(s)={D^t(s)}/{\mu^t(s)} < 1
\label{rho}
\end{equation}
is the stability constraints for queuing system in \cite{RN17,RN18}. When we set maximum $W(s)$ to $W_{\text{max}}$, and replace it with $W(s)$ into Eq. (\ref{mu}), (\ref{W(s)}) and (\ref{rho}), we can obtain the maximum $D_{\text{max}}^t$ through:
\begin{equation}
    D_{\text{max}}^t=\frac{2W_{\text{max}}\left(C^t(s)\right)^2}{\left(2W_{\text{max}}C^t(s)+B\right)B}.
\end{equation}
\par
Therefore, we estimate the corrected number of EVs approaching CS $s$ at $t$ as follows:
\begin{equation}
\tilde{D}^t(s) = \begin{cases}
    D_{\text{max}}^t, & \text{} D^t(s) \geq D_{\text{max}}^t \\
    D^t(s), & \text{} D^t(s) < D_{\text{max}}^t
    \end{cases},
\label{corrected D}
\end{equation}
and corrected $\tilde{W}^t(s)$ can be obtained by:
\begin{equation}
\tilde{W}^t(s) = \begin{cases}
    W_{\text{max}}, & \text{} D^t(s) \geq D_{\text{max}}^t \\
    \frac{ \rho ^t(s)}{2 \mu ^t(s)\left(1- \rho ^t(s)\right)}, & \text{} D^t(s) < D_{\text{max}}^t
    \end{cases},
\label{corrected D}
\end{equation}
And we estimate $charging\ time$ of charging plan $P$ through
\begin{equation}
charging(P)=\dfrac{1}{|T|}\sum_{t\in T}\sum_{s\in S}\dfrac{D^t(s)}{\mu^t(s)}.
\end{equation}

Finally, we aggregate the travel time, charging time, and waiting time for a charging plan ($P$) into a single cost function:
\begin{equation}
\begin{split}
cost(P) &= \alpha \cdot travel(P)\\& + (1 - \alpha) \cdot \left[waiting(P)+charging(P)\right],
\end{split}
\end{equation}
% \begin{equation}
%     \resizebox{.91\linewidth}{!}{$
%             \displaystyle
%            cost = \alpha \cdot travel(P) + (1 - \alpha) \cdot \left[waiting(P)+charging(P)\right]
%         $},
% \end{equation}
where \( \alpha \in [0,1] \) is a weighting parameter.

\subsubsection{Queuing Loss}

Since we have introduced the maximum waiting time $W_{\text{max}}$ and corrected estimated number of EVs $\tilde{D}^t(s)$ in Section \ref{sec: cost funtion}, a metric so-called $queuing\ loss$ is introduced to measure the loss of those demand for avoiding long waiting time, i.e., 
\begin{equation}
    queuing\ loss(P)=\sum_{t\in T}\sum_{s\in S}\max \{ D^t(s)-\tilde{D}^t(s) ,0\}.
\end{equation}

Finally, we model utility function by combining three evaluation metrics through:

% \begin{equation}
% \label{eq: utility}
%     \resizebox{.91\linewidth}{!}{$
%             \displaystyle
%            utility(P) = \lambda_b \cdot benefit(P)+\lambda_c\cdot cost(P)+\lambda_q\cdot queuing\ loss(P)
%         $},
% \end{equation}
\begin{equation}
\label{eq: utility}
\begin{split}
utility(P)&= \lambda_b \cdot benefit(P)+\lambda_c\cdot cost(P)\\&+\lambda_q\cdot queuing\ loss(P),
\end{split}
\end{equation}
where $\lambda_b$, $\lambda_c$ and $\lambda_q$ are trade-off weights.

\subsection{Mobile Chargers Scheduling}
We propose two heuristic strategies for scheduling mobile chargers (MCs): supporting overloaded charging stations and establishing flexible charging areas. Additionally, we introduce policies for MC adjustment and recall to adapt to dynamic demand.

\subsubsection{Supporting Charging Stations}
To support overloaded charging stations, we employ a heuristic strategy: (1) identify the station $s$ with the highest loss, calculated as $\lambda_c \cdot (waiting + charging) + \lambda_q \cdot queuing\ loss$; (2) sort all idle MCs by their distance to $s$; (3) select the nearest idle MC with sufficient energy ($> c_{MC}$) and not currently scheduled; (4) assign it to $s$ and update the station's service capacity and related metrics accordingly.

\subsubsection{Establishing Flexible Charging Areas}
% Beyond fixed CSs, HCSPO allows MCs to be relocated to any node without a CS, forming flexible charging areas. At each decision step, we identify the non-CS node $v$ with the highest cumulative demand $dem^{t:|T|}(v)$, then dispatch the nearest idle MC to it. A temporary charging site is created at $v$ without fixed infrastructure, and system metrics such as capacity $C$, service rate $\mu$, and adjusted demand $\tilde{D}$ are updated accordingly.

To supplement fixed charging stations, we employ a heuristic strategy:
(1) identify the non-CS node $v$ with the highest cumulative demand $dem^{t:|T|}(v)$;
(2) sort all idle MCs by their distance to $v$;
(3) select the nearest idle MC with sufficient energy ($> c_{MC}$) and not currently scheduled;
(4) dispatch it to $v$ to establish a temporary charging site without fixed infrastructure;
(5) update system metrics such as capacity $C$, service rate $\mu$, and adjusted demand $\tilde{D}$ accordingly.

\subsubsection{Adjustment and Recall Policy for MCs}
\label{sec: energy adjustment}

Relocating a MC $m$ incurs a scheduling delay during which it cannot provide service. The next arrival time after scheduling is given by $\tau^{t+1}_m = \text{dist}(l^t_m, l^{t+1}_m)/\text{vel}$. To account for the service loss during this period, we define a discount factor:
\begin{equation}
\delta^{t+1}_m = 1 - \min\left\{\frac{\tau^{t+1}_m - (t+1)H}{H}, 1\right\},
\end{equation}
where $H$ is the duration of a time slot.

Since MCs have limited battery capacity, their remaining energy is updated as:
\begin{equation}
q^{t+1}_m = q^t_m - \sum_{s \in S} i_{MC}(m, s, t) \cdot \max\{c_{MC} \delta^t_m, 0\}.
\end{equation}

If $q^t_m < c_{MC}$, the MC is recalled to a depot $j$ for recharging. The required energy is $q = B_{MC} - q^t_m$, and the updated arrival time becomes $\tau^{t+1}_m = \text{dist}(l^t_m, j)/\text{vel} + C'/q$, where $C'$ is the charging power.

\section{Reinforcement Learning Framework}

% In this section, we introduce an adaptive reinforcement learning framework for HCSPO problem (ARL-HCSPO) designed to optimize the placement of charging stations and configurations of fixed chargers with operation of Mobile chargers scheduling. This framework integrates the predictive model from Section \ref{sec: demand prediction} and uses the utility function described in Section \ref{sec: utility model} to construct a reward function, guiding decision-making in a dynamic environment.

We propose an adaptive reinforcement learning framework (\textsc{ARL-HCSPO}) to jointly optimize station placement, fixed charger setup, and mobile charger scheduling, using demand prediction (Section~\ref{sec: demand prediction}) and utility modeling (Section~\ref{sec: utility model}) to shape the reward in dynamic environments.

\noindent \textbf{Observation.} The observation $O^{i}$ at episode step $i$ consists of two components: planning observation for CSs, denoted as $O_{CS}^i$ and operation observation for MCs, denoted as $O^{i}_MC$. 
The planning observation is defined as follows $O_{CS}^i=\left\{\left(lon_v,lat_v,dem^{1,\cdots,|T|}_v,x \right)_v|\forall v \in V \right\}$, where $lon_v$ and $lat_v$ are the coordinate of node $v$, and $x=\left(x_1,\cdots,x_n \right)$ is the configuration of fixed chargers, which is only applicable when $v$ is a CS.

Then we define the operation observation using equation $O_{MC}^i=\left\{\left(l_m^{1,\cdots,|T|},q_m^{1,\cdots,|T|}, \tau_m^{1,\cdots,|T|} \right)_m|\forall m \in M \right\}$, which captures the location, arrival time after scheduling and remaining electricity of all time slots $t\in T$.

% \begin{equation}
%     O_{CS}^i=\left\{\left(lon_v,lat_v,dem^{1,\cdots,|T|}_v,x \right)_v|\forall v \in V \right\},
% \end{equation}
% where $lon_v$ and $lat_v$ are the coordinate of node $v$, and $x=\left(x_1,\cdots,x_n \right)$ is the configuration of fixed chargers, which is only applicable when $v$ is a CS.\par
% And we define operation observation $O^{i}_{MC}$ through
% \begin{equation}
%     O_{MC}^i=\left\{\left(l_m^{1,\cdots,|T|},q_m^{1,\cdots,|T|}, \tau_m^{1,\cdots,|T|} \right)_m|\forall m \in M \right\},
% \end{equation}
% which captures the location, arrival time after scheduling and remaining electricity of all time slots $t\in T$.

\noindent \textbf{Action.} We adopt a five-action space inspired by \cite{RN17}, serving as neighborhood operations to modify the charging plan:

\begin{itemize}[leftmargin=*]
\item \textit{Create by Demand}: Add a CS at the node with highest demand using a precomputed capacity-cost lookup table over $n$ charger types.
\item \textit{Create by Benefit}: Add a CS at the node with lowest coverage; configuration follows the same lookup rule.
\item \textit{Increase by Demand}: Add one charger to the node with highest demand.
\item \textit{Increase by Benefit}: Add one charger to a CS near the lowest-benefit node.
\item \textit{Relocate}: Move a charger from the lowest-benefit CS to the one with the highest unmet demand (e.g., waiting time or queuing loss).
\end{itemize}

To balance planning and real-time operation, after each action step $i$, we schedule mobile chargers (MCs) based on the updated plan. This coordination ensures flexibility for demand surges via MCs, while CSs provide stable long-term coverage.

\noindent \textbf{Reward.} Given the utility function described in Section \ref{sec: utility model}, we define the reward function between transition as the difference between $P^{i+1}$ and $P^{i}$, i.e., $r^i=utility(P^{i+1})-utility(P^{i})$.

\section{Experiments}
In this section, we will introduce how we perform our ARL-HCSPO method on a real-world application. 
\subsection{Datasets}
We evaluate our approach on the \textbf{road network} of Nanshan District, Shenzhen, China, comprising 1663 nodes and 2964 edges, similar to \cite{RN17}. \textbf{Existing charging station} locations are collected from \cite{RN16} and used as a baseline. Due to the lack of precise charging demand data at the road level, we use charging session data from \cite{RN16} for \textbf{charging demand} estimation, allocating node-level demand via inverse distance weighting from each node to the nearest CS.

% \subsection{Evaluation Metrics}
% Based on the utility function in Section \ref{sec: utility model}, we evaluate the effectiveness of the charging plan generated by our approach using several metrics: $benefit$, $cost$, $queuing\ loss$, $travel$, $charging$, and $waiting$.

\subsection{Evaluation Metric and Implementation}
Based on the utility function in Section \ref{sec: utility model}, we evaluate the effectiveness of the charging plan generated by our approach using several metrics: $benefit$, $cost$, $queuing\ loss$, $travel$, $charging$, and $waiting$.

In the utility model, we use the following parameters as default setting: $\alpha=0.4$, $\lambda_b=0.4$, $\lambda_c=0.4$, $\lambda_q=0.2$, $n=3$, $k=25$, $Budget=5.4e7$ CNY, $E=35$ kwh, $r_{\text{max}} = 2.5$ km, $H=60$ min, $T$=3, $W_\text{max}=30$ min, $|M|=30$, $k_{MC}=10$, $c_{\text{MC}}=42$ kw, $B_{\text{MC}}=105$ kwh. The rated charging power of fixed chargers, $c_1$, $c_2$ and $c_3$ are set to 7kw, 22kw and 50kw, respectively, with corresponding installation fees of 2700 CNY, 6750 CNY and 252000 CNY. The operation cost for each MC is 59400 CNY.  

RL in this paper is applied on Stable Baselines 3 \cite{stable-baselines3}, implemented using DQN as basic model to train an optimal policy. We set learning rate to 0.01, buffer size to 10000, batch size to 128. The maximum number of training steps is set to 30000. 

\subsection{Baselines}

\begin{table*}[t]
% \vspace{-1em}
\centering
\begin{tabular}{l|c|c|c|c|c|c}
% \hline
% \multicolumn{8}{c}{Header} \\
    % \cline{1-8} 
    \noalign{\hrule height 1pt}
    % \toprule
     Approach & $benefit\ \uparrow$ & $cost \downarrow$ & $travel \downarrow$ & $charging \downarrow$ & $waiting \downarrow$ & $queuing\ loss \downarrow$\\
    % \midrule
    \noalign{\hrule height 0.5pt}
    \textsc{Existing Plan} & 100\% & 100\% & 100\% & 100\% & 100\% & 100\% \\
    \textsc{Highest Demand} & 93\% & 90\% & 91.9\% & 90.4\% & 72.7\% & 98.2\% \\
    \textsc{Bounding\&Optimizing+} & 78\% & 97.3\% & 98.5\% & 123.2\% & 73.2\% & 95.7\% \\
    \textsc{PCRL} & 126\% & 63.8\% & 71.1\% & 44.9\% & 21\% & 94.2\% \\
    \textsc{ARL-HCSPO} (ours) & \textbf{205.1\%} & \textbf{59.5\%} & \textbf{52.8\%} & \textbf{43.3\%} & \textbf{20.2\%} & \textbf{86.7\%} \\
   % \bottomrule
    \noalign{\hrule height 1pt}
\end{tabular}
\caption{Results on Nanshan, Shenzhen dataset compared to baselines. Evaluation metrics where higher value are better are marked with {\textbf{$\uparrow$}}, otherwise are marked with {\textbf{$\downarrow$}}. Best results are marked bold.}
\label{tab:metrics}
% \vspace{-1em}
\end{table*}

\begin{table*}[t]
% \vspace{-1em}
\centering
\begin{tabular}{l|c|c|c|c|c|c}
% \hline
% \multicolumn{8}{c}{Header} \\
    % \cline{1-8} 
    \noalign{\hrule height 1pt}
    % \toprule
     Approach & $benefit\ \uparrow$ & $cost \downarrow$ & $travel \downarrow$ & $charging \downarrow$ & $waiting \downarrow$ & $queuing\ loss \downarrow$\\
    % \midrule
    \noalign{\hrule height 0.5pt}
    \textsc{Existing Plan} & 100\% & 100\% & 100\% & 100\% & 100\% & 100\% \\
    \textsc{ARL-HCSPO (w/o MPC)} & 122.8\% & 62.7\% & 68.6\% & 45.4\% & 20.2\% & 91.9\% \\
    \textsc{ARL-HCSPO (w/o MCS1)} & \textbf{205.7\%} & 63.9\% & 70.2\% & \textbf{42.6\%} & 26.2\% & 93\% \\
    \textsc{ARL-HCSPO (w/o MCS2)} & 141.3\% & \textbf{48.5\%} & \textbf{29.1\%} & 45.3\% & \textbf{11.9\%} & \textbf{67.8\%} \\
    \textsc{ARL-HCSPO} & 205.1\% & 59.5\% & 52.8\% & 43.3\% & 20.2\% & 86.7\% \\
   % \bottomrule
    \noalign{\hrule height 1pt}
\end{tabular}
\caption{Ablation study on Nanshan, Shenzhen dataset. Evaluation metrics where higher value are better are marked with {\textbf{$\uparrow$}}, otherwise are marked with {\textbf{$\downarrow$}}. Best results are marked bold.}
\label{tab:ablation}
% \vspace{-1em}
\end{table*}

% To evaluate the effectiveness of our approach (denoted as \textsc{ARL-HCSPO}), we compare it against several state-of-the-art baselines, detailed below, using predefined evaluation metrics. Each method is executed with the maximum budget as the default setting. It is worth noting that the baselines do not incorporate mobile chargers; therefore, for fairness, these methods are allowed to install more fixed chargers within the same budget constraints, as specified in Eq. \ref{eq: budget limitation}.

% \begin{itemize}[leftmargin=*]
% \item \textbf{\textsc{Existing Charging}}:  We introduce existing placement of location-fixed charging station as the basic baselines. By comparing metrics with it, we can straightforwardly quantify how much the charging plan is improved by our approach.
% \item \textbf{\textsc{Highest Demand}}:  We introduce an greedy algorithm as \textsc{Highest Demand} by greedily selecting the road nodes with highest sum demand until budget is exhausted. The configurations of fixed chargers are determined the same as discussed in action space.  
% \item \textbf{\textsc{Bounding\&Optimizing+}}: This method is introduced as a baseline in \cite{RN17}, which improves method of \cite{RN18} by introducing the configuration lookup table method as a greedy strategy.
% \item \textbf{\textsc{PCRL}}: This is employed to address the real-world problem of placing location-fixed charging stations \cite{RN17}, which we introduce as a baseline in this paper. 
% \end{itemize}

To evaluate the effectiveness of our method (\textsc{ARL-HCSPO}), we compare it with several state-of-the-art baselines under the same budget constraint (Eq.~\ref{eq: budget limitation}). Since baselines do not utilize mobile chargers, they are allowed to deploy more fixed chargers for fairness.

\begin{itemize}[leftmargin=*]
\item \textbf{\textsc{Existing Plan}}: Uses the real-world deployment of location-fixed charging stations as a reference to quantify improvement.
\item \textbf{\textsc{Highest Demand}}: A greedy strategy that selects nodes with the highest cumulative demand until the budget is depleted; charger configurations follow our action space setup.
\item \textbf{\textsc{Bounding\&Optimizing+}}: Based on \cite{RN17}, this method enhances \cite{RN18} by integrating a configuration lookup table for greedy allocation.
\item \textbf{\textsc{PCRL}}: A reinforcement learning-based approach from \cite{RN17} for optimizing fixed CS placement in real-world scenarios.
\end{itemize}

\subsection{Experimental Results}
In this experiment we apply our approach as well as other baselines to solve HCSPO problem on Nanshan district, Shenzhen datasets. 

\subsubsection{Evaluation}

Table \ref{tab:metrics} presents the performance metrics of our approach compared to other baselines, where \textsc{ARL-HCSPO} consistently outperforms across all metrics. Notably, the $benefit$ metric sees a significant improvement at 205.1\%, vastly exceeding other methods, indicating that our approach can substantially enhance the availability of charging infrastructure across the road network. Additionally, a 40.5\% reduction in the $cost$ metric demonstrates our approach's ability to improve EV owners' charging satisfaction, especially reflected in $waiting$ metric with a 79.8\% reduction.

\begin{figure}[!t]
    \centering
    \includegraphics[width=0.95\linewidth]{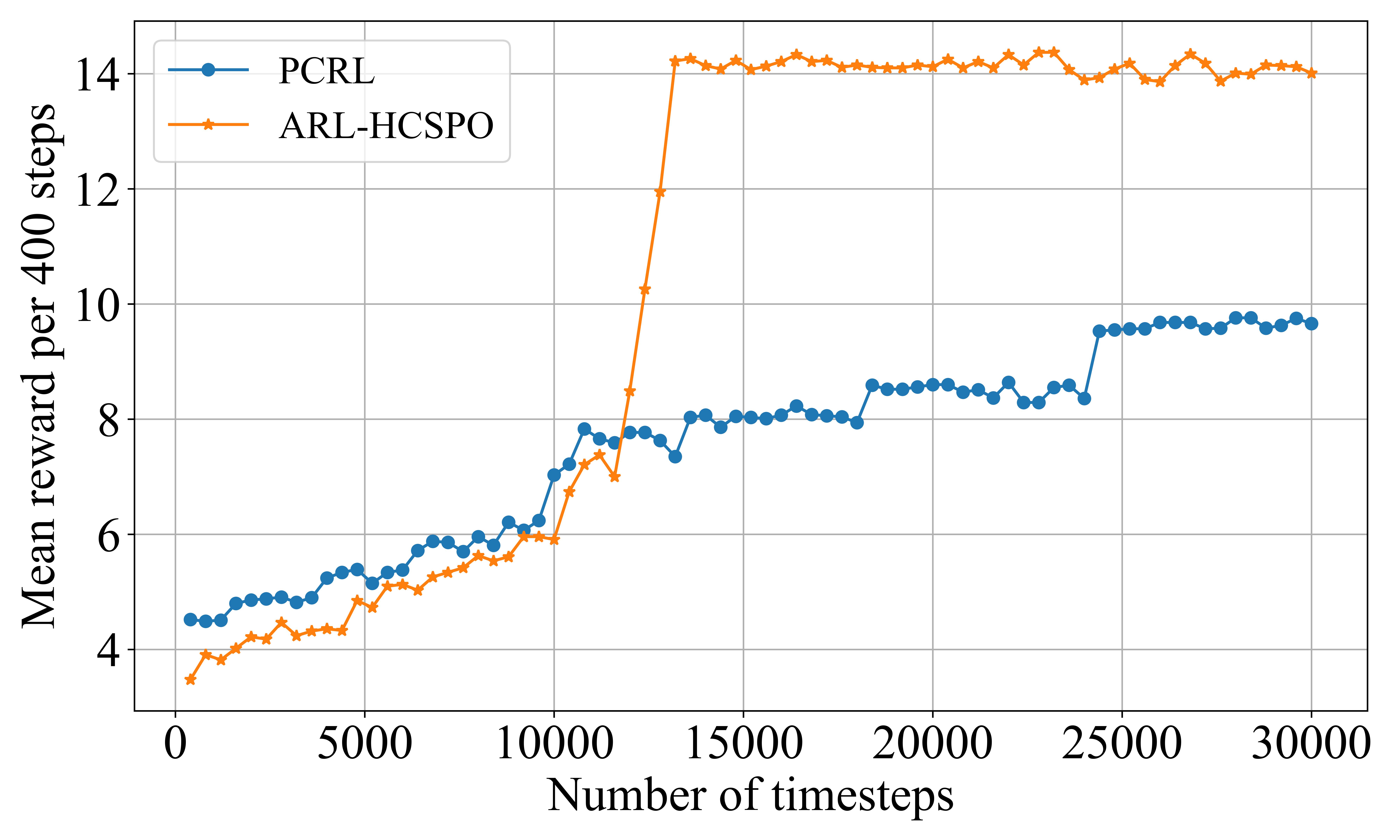}
    % \vspace{-1em}
    % \vspace{-1em}
    \caption{Training progression of \textsc{ARL-HCSPO} and \textsc{PCRL}, with mean rewards evaluated every 400 training steps.}
    \label{fig:reward}
    % \vspace{-1em}
\end{figure}

For RL-based approach, i.e., \textsc{PCRL} and \textsc{ARL-HCSPO}, we initialize charging plan as empty before training. Figure \ref{fig:reward} illustrates the training evolution of our approach compared to \textsc{PCRL}. As shown, our method (in blue) outperforms \textsc{PCRL} (in orange) by achieving higher episode rewards with more stable convergence.

\begin{figure}[!t]
    \centering
    \includegraphics[width=1\linewidth]{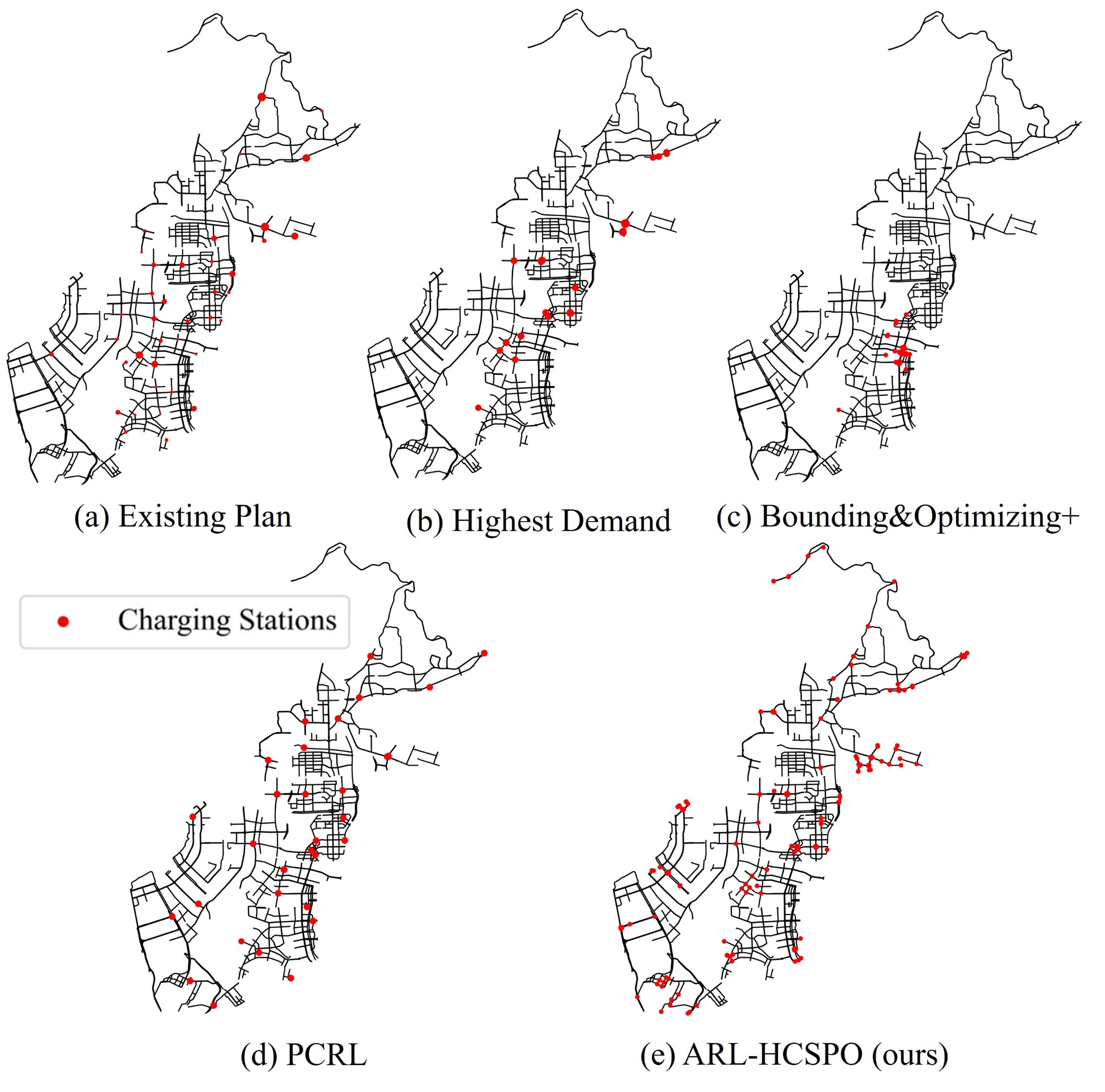}
    % \vspace{-1em}
    % \vspace{-1em}
    \caption{Charging plan comparison on Nanshan district, Shenzhen.}
    \label{fig:comparison of charging station planning distribution}
    % \vspace{-1em}
\end{figure}

Figure \ref{fig:comparison of charging station planning distribution} depicts the distribution of charging stations generated by various methods. On this dataset, both the \textsc{Bounding\$Optimizing+} and \textsc{High Demand} approaches fail to show significant improvement over the \textsc{Existing Plan}. In contrast, both \textsc{PCRL} and \textsc{ARL-HCSPO} exhibit noticeable enhancements, with \textsc{ARL-HCSPO} achieving the most extensive CS coverage, due to the integration of mobile charger scheduling. Moreover, \textsc{ARL-HCSPO} results in a more evenly distributed network of CSs across the road system.

% \vspace{-1em}

\subsubsection{Ablation Study}
To evaluate the effectiveness of each module within \textsc{ARL-HCSPO}, we compare our approach against three variants: (1) \textsc{ARL-HCSPO (w/o MPC)}, which excludes the MPC policy; (2) \textsc{ARL-HCSPO (w/o MCS1)}, which removes the supporting charging station scheduling strategy; and (3) \textsc{ARL-HCSPO (w/o MCS2)}, which omits the flexible charging areas scheduling strategy. 

The results in Table \ref{tab:ablation} indicate that all variants improve upon the \textsc{Existing Plan}. Comparing our approach with \textsc{ARL-HCSPO (w/o MPC)}, we observe a significant 82.3\% increase in the $benefit$ metric, alongside modest gains in other metrics, underscoring the value of incorporating the demand prediction model and MPC policy. Additionally, the analysis of the two MC scheduling strategies reveals different focal points: \textsc{ARL-HCSPO (w/o MCS1)} emphasizes maximizing $benefit$, achieving a peak $benefit$ of 205.7\%, slightly exceeding \textsc{ARL-HCSPO} (205.1\%). Conversely, \textsc{ARL-HCSPO (w/o MCS2)} prioritizes cost-efficiency, recording the lowest $cost$ at 48.5\%, with significant reductions in $travel$ (29.1\%) and $waiting$ (11.9\%). 

The results from \textsc{ARL-HCSPO} demonstrate that a balanced approach, combining both strategies, can deliver a substantial $benefit$ increase of 63.8\% with only an 11\% increase in $cost$ compared to \textsc{ARL-HCSPO (w/o MCS2)}.

\subsubsection{Effect of $K$}
\begin{figure}[t]
    \centering
    \includegraphics[width=\linewidth]{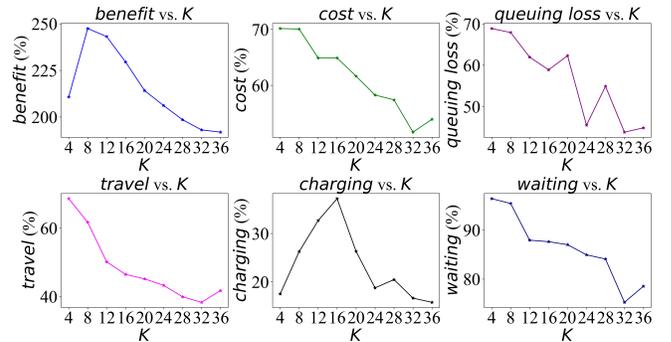}
    % \vspace{-1em}
    % \vspace{-1em}
    \caption{Impact of various metrics compared to \textsc{Existing Plan} as percentages, evaluated against different values of $K$.}
    \label{fig:sensitivity for k}
    % \vspace{-1em}
\end{figure}

The maximum number of fixed chargers, $K$, is a critical factor in determining the effectiveness of our approach. Figure \ref{fig:sensitivity for k} illustrates the impact of varying $K$ values on different performance metrics when applying \textsc{ARL-HCSPO}. The results for all metrics are normalized by the scaled-down value of the \textsc{Existing Plan}, as shown in Table \ref{tab:metrics}.

For the $benefit$ metric, increasing $K$ from 4 to 36 results in an optimal value of 244.4\% at $K=8$, after which there is a moderate decline. In contrast, metrics such as $cost$, $travel$, and $waiting$ continue to improve as $K$ increases, reaching their optimal values of 51.6\%, 38.3\%, and 75.2\%, respectively, at $K=32$. The $queuing\ loss$ metric exhibits some fluctuation but generally trends downward.

In summary, while increasing $K$ improves performance in terms of $cost$ and $queuing\ loss$, it is important to strike a balance by keeping $K$ within a smaller range to optimize $benefit$. This balance is crucial for improving the overall supply of charging infrastructure.

Further sensitivity analysis could provide valuable managerial insights for future applications.

% {\fontsize{7pt}{15pt}\selectfont Test words 12}
% \vspace{-1em}

\section{Conclusion}
% In conclusion, this paper presents a comprehensive solution to the HCSPO problem by integrating the strategic planning of fixed charging stations with the dynamic operation of mobile chargers. We incorporate a demand prediction model and design two heuristic scheduling strategies for mobile chargers. An adaptive reinforcement learning algorithm, enhanced by heuristics, is used to solve the HCSPO. Extensive experiments with real-world datasets demonstrate that our method significantly outperforms existing baseline approaches across various metrics, providing an efficient framework to address the challenges of dynamic EV charging demand.

In conclusion, this paper presents a comprehensive solution to the HCSPO problem by integrating fixed charging station planning with dynamic mobile charger operation. We propose a demand prediction model and two heuristic scheduling strategies for mobile chargers, solved using an adaptive reinforcement learning algorithm. Extensive experiments on real-world datasets show that our method outperforms existing approaches across multiple metrics, offering an efficient solution to dynamic EV charging demand challenges.

\section*{Acknowledgments}
This work is mainly supported by the Guangdong Basic and Applied Basic Research Foundation (No. 2025A1515011994). This work is also supported by  the National Natural Science Foundation of China (No. 62402414), Guangzhou Municipal Science and Technology Project (No. 2023A03J0011), the Guangzhou Industrial Information and Intelligent Key Laboratory Project (No. 2024A03J0628), and a grant from State Key Laboratory of Resources and Environmental Information System, and Guangdong Provincial Key Lab of Integrated Communication, Sensing and Computation for Ubiquitous Internet of Things (No. 2023B1212010007).

\bibliographystyle{named}
\bibliography{ref}

\end{document}